\documentclass[10pt,journal,compsoc]{IEEEtran}
\ifCLASSOPTIONcompsoc
  \usepackage[nocompress]{cite}
\else
  \usepackage{cite}
\fi
\ifCLASSINFOpdf
\else
\fi
\usepackage{graphicx}
\usepackage{subfig}
\usepackage{float}
\usepackage{enumitem}
\usepackage{hyperref}

\begin{document}
\title{DAiSEE: Towards User Engagement Recognition in the Wild}

\author{Abhay~Gupta, Arjun~D'Cunha, Kamal~Awasthi and~Vineeth~Balasubramanian% <-this % stops a space
\IEEEcompsocitemizethanks{\IEEEcompsocthanksitem A. Gupta~abhgup@microsoft.com
\IEEEcompsocthanksitem A. D'Cunha~cs14btech11039@iith.ac.in
\IEEEcompsocthanksitem K. Awasthi~201451072@iiitvadodara.ac.in
\IEEEcompsocthanksitem V. Balasubramanian~vineethnb@iith.ac.in}% <-this % stops a space
\thanks{}}


\markboth{Journal of \LaTeX\ Class Files,~Vol.~14, No.~8, August~2015}%
{Shell \MakeLowercase{\textit{et al.}}: Bare Advanced Demo of IEEEtran.cls for IEEE Computer Society Journals}



\IEEEtitleabstractindextext{%
\begin{abstract}
The difference between real and virtual worlds is shrinking at an astounding pace. With more and more users working on computers to perform a myriad of tasks from online learning to shopping, interaction with such systems is an integral part of life. In such cases, recognizing a user's engagement level with the system (s)he is interacting with can change the way the system interacts back with the user. This will lead not only to better engagement with the system but also pave the way for better human-computer interaction. Hence, recognizing user engagement can play a crucial role in several contemporary vision applications including advertising, healthcare, autonomous vehicles, and e-learning. However, the lack of any publicly available dataset to recognize user engagement severely limits the development of methodologies that can address this problem. To facilitate this, we introduce DAiSEE, the first multi-label video classification dataset comprising of 9068 video snippets captured from 112 users for recognizing the user affective states of boredom, confusion, engagement, and frustration ``in the wild''. The dataset has four levels of labels namely - very low, low, high, and very high for each of the affective states, which are crowd annotated and correlated with a gold standard annotation created using a team of expert psychologists. We have also established benchmark results on this dataset using state-of-the-art video classification methods that are available today. We believe that DAiSEE will provide the research community with challenges in feature extraction, context-based inference, and development of suitable machine learning methods for related tasks, thus providing a springboard for further research.
\end{abstract}

\begin{IEEEkeywords}
Affect Recognition, User Engagement in the Wild, DAiSEE, E-Learning Environments
\end{IEEEkeywords}}

\maketitle

% is not selected <OR> if conference mode is selected - because compsoc
% conference papers position the abstract like regular (non-compsoc)
\IEEEdisplaynontitleabstractindextext

% \fi
\IEEEpeerreviewmaketitle

\ifCLASSOPTIONcompsoc
\IEEEraisesectionheading{\section{Introduction}\label{sec:introduction}}
\else
\section{Introduction}
\label{sec:introduction}
\fi

\IEEEPARstart{T}{he} progress from research to consumer technologies in classical recognition problems in computer vision have been possible over the last decade due to the availability of large-scale datasets which are made available to researchers and industry practitioners alike. The ImageNet \cite{deng2009imagenet} and PASCAL VOC \cite{everingham2010pascal} challenges spearheaded the object recognition revolution, while the availability of datasets such as Microsoft COCO \cite{lin2014microsoft} and Cityscapes \cite{cordts2016cityscapes} in recent years has fueled the development of newer methods for semantic segmentation and vision-language joint understanding. Needless to say, the availability of more datasets for various problems and subproblems in computer vision will allow for a greater impact in translating research methodologies to businesses and, eventually, a positive influence on users' lives. This work is an effort in this direction - to provide a dataset (and benchmark results) for a vision problem of contemporary relevance: user engagement recognition.

\begin{figure}
\includegraphics[width=85mm, height=100mm]{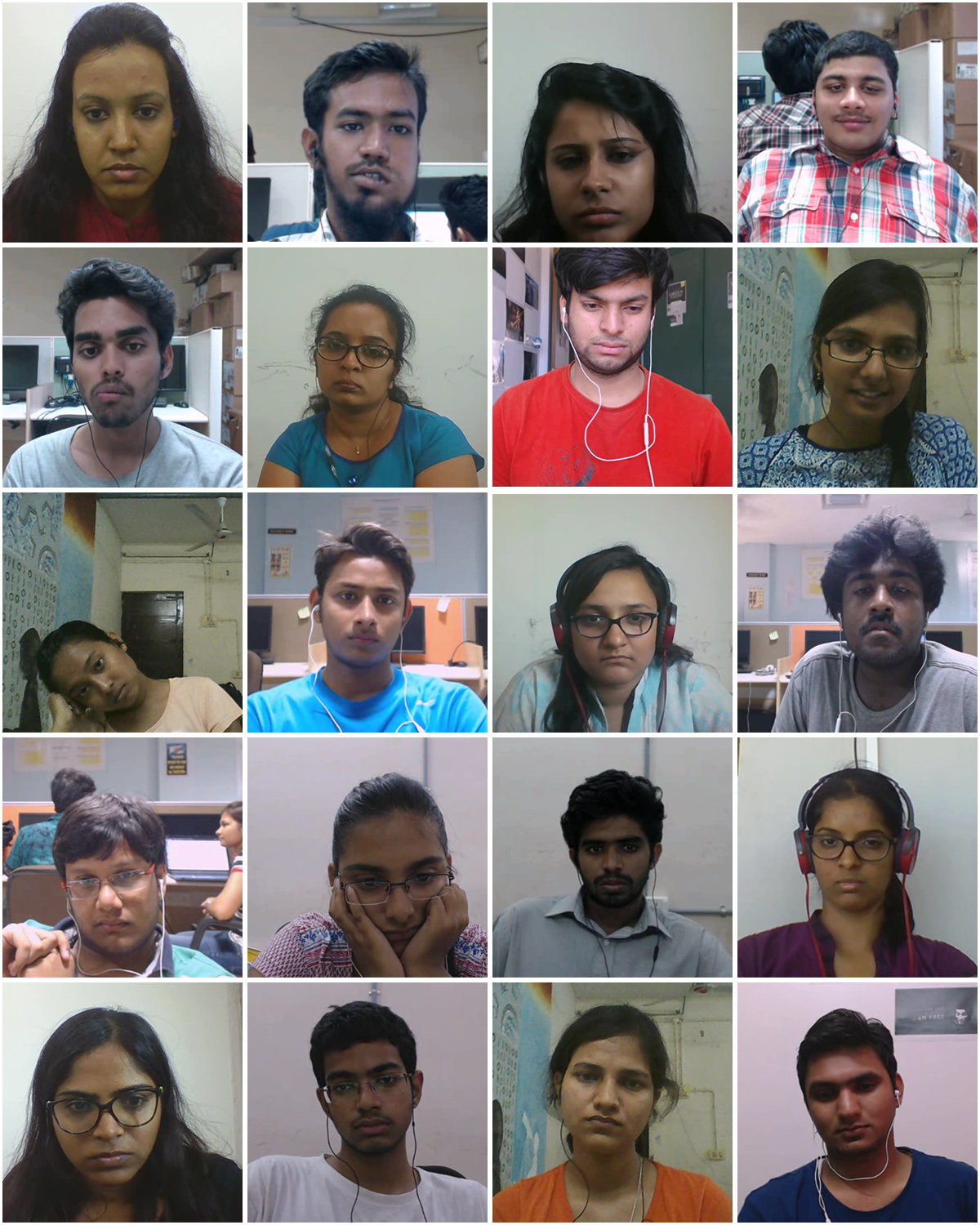}
\caption{Examples of video frames from DAiSEE: The dataset captures real-world challenges of recognizing user engagement in natural settings.}
\label{fig_daisee}
\end{figure}

The recognition of user engagement is increasingly relevant to a digital world that floods users with various kinds of content, and it is useful for systems to be ``aware" of the user's engagement while providing content. For example, the recognition of user engagement is critical to various domains including advertising - web or television (is the viewer really watching this advertisement on television or on YouTube? is (s)he bored?); healthcare - as pertains to detection and interventions for children with learning or cognitive disabilities (does a child's engagement levels indicate a tendency for autism or ADHD? are there particular events or objects that interest or frustrate the child?); e-learning (which parts of a lecture are confusing for most students who watch it? how engaged are students in a video?); and autonomous vehicles (Is the driver distracted or is he paying attention? Can we predict his actions based on his confusion/frustration levels?). There have been recent research efforts to develop methods for user engagement recognition \cite{whitehill2014faces}\cite{hernandez2013measuring}, but the resulting datasets are very small and not publicly available. Further, existing commercially available affective recognition systems that attempt to track user engagement work within constrained settings and have limited use in real-world environments (illustrated in Section~\ref{sec_background}). This work introduces DAiSEE, a dataset that aims to facilitate research and development towards user engagement recognition in the wild.

Understanding affective states of a user, an important subarea of computer vision, has for a long time focused on datasets pertaining to the seven basic expressions: neutral, happiness, sadness, anger, disgust, surprise and contempt \cite{lucey2010extended}. While recent efforts in the last few years have expanded datasets in this domain to cover affective states in terms of dimensional representations \cite{schuller2012avec}\cite{rehg2013decoding}, the vast subtleties in affective states necessitate the development of datasets for specific objectives. Recent trends, including \cite{valstar2014avec}\cite{rehg2013decoding}, corroborate this approach to help progress towards tangible outcomes.

E-learning environments provide one of the best use cases for studying user engagement in settings where a user interacts with a computer screen. With the accelerated growth of Massive Open Online Courses (MOOCs), there is a need to design intelligent interfaces which simulate the interactions that occur between a teacher and students in a class. The main drawback of existing e-learning systems is that they do not provide real-time interactive feedback to students (or instructors) during the content delivery process, as compared to traditional classroom learning. Surprisingly, MOOCs have a dropout rate of 91-93\%~\cite{jordan2014initial} \cite{mooc@drop}, with the completion rate for the first assignment being around 45\%. An online survey~\cite{colman2013mooc} lists the top ten reasons for dropouts from such platforms to include: ``poor course design'', which included components such as lack of proper feedback, ``lecture fatigue" in courses that had only video lectures, ``lack of proper course introductions'' and ``student frustration''. Such reasons motivate the need to improve feedback mechanisms to make these platforms more interactive. Understanding user engagement at various junctures of the e-learning experience can help design intuitive interfaces that support better knowledge absorption by students, help decrease dropout rates, and personalize the learning experience. This paper seeks to address the aforementioned issues, by making available a dataset that captures user engagement in the wild during e-learning sessions (which is relevant to any other application domain involving user interaction with a computer screen including advertising, healthcare, and autonomous vehicles).

In this work, we introduce DAiSEE (\textbf{D}ataset for \textbf{A}ffect\textbf{i}ve \textbf{S}tates in \textbf{E}-\textbf{E}nvironments) (Figure~\ref{fig_daisee} shows sample images), where e-environment means any environment with applications like e-shopping, e-healthcare, e-learning but not limited to them. DAiSEE will be made publicly available to the community for further research. In particular, we focus on engagement, frustration, confusion, and boredom as the affective states for this work, all of which are relevant to user engagement and concomitant applications. Considering that these affective states are subtle, the annotations for this dataset are crowd-sourced, strengthened using a gold standard created using expert psychologists and all annotations (including each individual crowd annotation) are provided along with this dataset. We also benchmark the performance of standard video classification and deep learning-based models on this dataset to provide a baseline for further research. DAiSEE has the potential to be applied to the following domains, but is not limited to them:

\begin{itemize}
  \item In e-learning, to support personalized learning for an individual user, and thus, increase retention rate.
  \item In advertising, to gain insights into what piques a customer's interest, to see how engaging an advertisement is, and to then provide personalized advertising to customers.
  \item In e-shopping, to understand user preferences in terms of specific items, or a larger domain (clothing, jewelry, electronics, etc.) of interest to a user, and thus allow personalization in the shopping experience.
  \item In healthcare, to understand and detect early signs of ADHD or autism, or to study fatigue in radiologists viewing images on a screen.
  \item In autonomous vehicles, to capture the driver's engagement level, or to predict the driver's future actions based on his/her confusion or frustration levels.
\end{itemize}

The rest of the paper is organized as follows: we discuss the background and related work in Section~\ref{sec_background}. In Section~\ref{sec_daisee_dataset}, we introduce the dataset, its components, and details of the process involved in collecting it. In Section~\ref{sec_benchmark}, we provide a baseline performance on all labels in this dataset using recent state-of-the-art deep learning models for video classification. Section~\ref{sec_benchmarking_challenges} discusses the challenges that DAiSEE poses in establishing a benchmark for the dataset and Section~\ref{sec_release} discusses the release and distribution of the dataset. Lastly, in Section \ref{sec_conc}, we summarize our work and suggest future directions with our dataset.

\section{Background and Related Work}
\label{sec_background}

Determining the affective state of a user using computer vision and machine learning methods has been studied for over two decades now (please see \cite{zeng2009survey}\cite{tao2005affective} for surveys).

\subsection{Affect Recognition Datasets} 
Early databases of facial expressions such as CK+~\cite{tian2001recognizing}\cite{lucey2010extended}, MMI \cite{pantic2005web}, and MultiPie \cite{gross2010multi} were captured in a lab-controlled environment where the subjects portrayed different facial expressions, which resulted in a clean and high-quality dataset of posed facial expressions. However, posed expressions may differ from real-life spontaneous expressions. Thus, datasets like DISFA \cite{mavadati2013disfa}, AM-FED \cite{mcduff2013affectiva} which captured facial responses to stimuli or Belfast \cite{sneddon2012belfast} which captured emotions while performing laboratory-based tasks became a trend in the affective computing community. These datasets captured multi-model affects such as voice, biological signals and often worked with a series of frames to allow for both temporal and dynamic expression recognition. However, the diversity of the datasets is limited due to the number of subjects, head pose variation, and environmental conditions. 

To develop systems that are based on natural, unposed facial expressions - datasets in the wild became important. The \textit{Acted Facial Expressions in the Wild (AFEW)} released by Dhall et al.\cite{dhall2013emotion} addressed the issue of temporal facial expressions by capturing videos of subjects. The dataset contains 330 subjects aged 1-77 years and is annotated with six basic expressions along with neutral. A static subset (SFEW) \cite{dhall2011static} of 700 images and 95 subjects was created by selecting some frames from AFEW and it contains unconstrained facial expressions, different head poses and close to real-world scenarios like occlusions and unconstrained illumination. 

The \textit{Facial Expression Recognition 2013 (FER-2013)} dataset \cite{goodfellow2013challenges} was created using the Google Search API that matched a set of 184 emotion-related keywords to capture the six basic expressions and the neutral expression. The dataset contains 35,887 images, most of which are in the wild. The images are grayscale, 48x48 pixels and have been cropped to retain only the face and the displayed emotions. FER-2013 is currently the biggest publicly available facial expression database in the wild settings. However, the resolution and quality of the images make it difficult for facial landmark detectors to extract landmarks and only categorical models of affect are provided with FER-2013. 

The \textit{Affective-MIT Facial Expression Dataset (AM-FED)} dataset \cite{mcduff2013affectiva} contains 242 facial videos (160K frames) of people watching commercials using their webcam. The recordings were taken in ``wild'' settings with varied contrast and illumination. The dataset has a frame-by-frame annotation for 14 FACS action units, head movements, and automatically detected landmark points. However, there is not much variation in head pose and the dataset has limited number of subjects. 

The FER-Wild \cite{mollahosseini2016facial} dataset contains 24,000 images which match emotion-related terms queried from search engines. The dataset contains annotations for the six basic expressions and neutral from two human labelers. Compared to FER-2013~\cite{goodfellow2013challenges}, FER-Wild only has higher resolution with facial landmarks points. It still does not address limitations like categorical modeling of affect and portraying more expressions.

EmotioNet \cite{benitez2016emotionet} consists of one million images of facial expressions downloaded from the Internet by selecting all words derived from the word ``feeling'' in WordNet \cite{miller1995wordnet}. The images are automatically annotated with AUs and AU intensities after a face detector \cite{viola2004robust} was used to detect faces. The images are labeled one of 23 (basic or compound) emotion categories defined in \cite{du2014compound} based on AUs. Experienced coders were used to annotate 100,000 images manually with AUs. EmotioNet is a ``in-the-wild'' dataset with a large amount of subject variation and is used to study the FACS model. However, it lacks the dimensional model of affect, and the emotion categories are not manually labeled, rather, defined automatically based on annotated AUs. 

Most of the datasets discussed above lacked the dimensional model of affect and were not in the continuous domain. Some datasets like Belfast \cite{sneddon2012belfast}, RELOCA \cite{ringeval2013introducing}, Aff-Wild Database \cite{zafeiriou2016facial} and AffectNet \cite{mollahosseini2017affectnet} are some examples of these datasets. These datasets are limited in number since the annotation of continuous dimensions is more expensive and requires trained annotators.

The \textit{RELOCA} \cite{ringeval2013introducing} dataset consists of 46 participants that participated in a video conference that required collaboration for completing a task. Multi-modal data such as audio, video, ECG and EDA were collected and six annotators measured arousal and valence. The dataset was one of the first attempts to model dimensional model of affect with multiple cues and modalities. However, it had only 46 participants and the videos were captured in a lab controlled experiment. 

The \textit{Aff-Wild} dataset \cite{zafeiriou2016facial} is the largest dataset that measures continuous affect in the valence-arousal space ``in-the-wild''. The dataset consists of 500 videos taken from YouTube displaying emotions while watching videos, performing activities and reacting to jokes and three annotators gave frame-by-frame annotations for valence and arousal. While the dataset modeled the temporal variance of affect, it is limited because of 500 subjects.

The \textit{AffectNet} dataset \cite{mollahosseini2017affectnet} is currently the largest database of categorical and dimensional models of affect in the wild. The dataset has 1,000,000 images queried from the web using emotion-related keywords with words corresponding to gender, age, and ethnicity.  The dataset has labels for the six basic expressions and it contains a ``None'' type that is used for expressions like bored, tired, confused, focused etc. The annotations are focused on the correctness of the affect type rather than intensity. The images are also annotated for valence and arousal in the continuous domain and the dataset contains 450,000 subjects. 

The Belfast Natural Induced Emotion Database \cite{sneddon2012belfast} provides examples of mild to moderately strong natural emotions in response to a series of laboratory-induced tasks. Each example is of varied length (between 5 seconds to 60 seconds in length) and the examples are labeled by information on self-report of emotion, the gender of the subject, and valence in the continuous domain. The dataset contains a total 1400 videos split into three sets of 570, 650 and 180 videos respectively. Of this, the first set contains labels for frustration, disgust, surprise, fear, and amusement. A task was designed to induce frustration/irritation/annoyance and the first 30 seconds of the subject's emotions were recorded. While natural emotions are portrayed in the dataset, an artificial setting of a laboratory is used where illumination, head-poses are controlled.

We can see that no dataset addresses the four affective states of boredom, confusion, engagement and frustration which are needed for applications such as e-learning, e-advertising etc. While the Belfast \cite{sneddon2012belfast} dataset does address the affective state of frustration, it is not ``in-the-wild'' and is stimuli-driven. Also, while AffectNet~\cite{mollahosseini2017affectnet} does try to address focus, boredom, confusion, it combines them into a single ``None'' class which is also not intensity annotated, not allowing for practical applications in the above-mentioned systems. 

A comparison of DAiSEE with several state-of-the-art datasets in the domain of facial expression/affective state recognition is presented in Table~\ref{tab_daisee_comparison}.

\begin{table*}[h]
\centering
\caption{Summary and Characteristics of Several Datasets in Affect Recognition}
\label{tab_daisee_comparison}
\resizebox{\textwidth}{!}{%
\begin{tabular}{|c|l|c|l|l|}
\hline
\textbf{Database} & \multicolumn{1}{c|}{\textbf{Database Information}} & \textbf{\# of Subjects} & \multicolumn{1}{c|}{\textbf{Condition}} & \multicolumn{1}{c|}{\textbf{Affect Modelling}} \\ \hline
FERG DB~\cite{aneja2016modeling} & \begin{tabular}[c]{@{}l@{}}- 55,767 images\\ - Frontal view\end{tabular} & 6 & - Posed & - 7 emotion categories \\ \hline
Oulu-CASIA-NIR-VIS~\cite{zhao2011facial} & \begin{tabular}[c]{@{}l@{}}- 2,880 videos\\ - Frontal view\end{tabular} & 80 & - Posed & - 7 emotion categories \\ \hline
AR Face Database~\cite{martinez1998ar} & \begin{tabular}[c]{@{}l@{}}- 4,004 images\\ - Frontal view\end{tabular} & 154 & - Posed & - 7 emotion categories \\ \hline
JAFFE Database~\cite{lyons1998coding} & - 213 images & 10 & - Posed & - 7 emotion categories \\ \hline
CK+~\cite{lucey2010extended} & \begin{tabular}[c]{@{}l@{}}- 593 images\\ - Frontal \& 30 degree images\end{tabular} & 123 & \begin{tabular}[c]{@{}l@{}}- Controlled\\ - Posed\end{tabular} & \begin{tabular}[c]{@{}l@{}}- 30 AUs\\ - 7 emotion categories\end{tabular} \\ \hline
MultiPie~\cite{gross2010multi} & \begin{tabular}[c]{@{}l@{}}- 750,000 images\\ - Multipl viewpoints \& illuminations\end{tabular} & 337 & \begin{tabular}[c]{@{}l@{}}- Controlled\\ - Posed\end{tabular} & - 7 emotion categories \\ \hline
MMI~\cite{pantic2005web} & \begin{tabular}[c]{@{}l@{}}- 2900 videos\\ \\ - Frontal and side views\end{tabular} & 25 & \begin{tabular}[c]{@{}l@{}}- Controlled\\ - Posed\\ - Spontaneous\end{tabular} & \begin{tabular}[c]{@{}l@{}}- 31 AUs\\ - 6 Basic Expressions\end{tabular} \\ \hline
DISFA~\cite{mavadati2013disfa} & \begin{tabular}[c]{@{}l@{}}- 130,000 video frames\\ - Frontal views\end{tabular} & 27 & \begin{tabular}[c]{@{}l@{}}- Controlled\\ - Spontaneous\end{tabular} & - 12 AUs \\ \hline
SALDB~\cite{nicolaou2010audio}~\cite{nicolaou2011continuous} & \begin{tabular}[c]{@{}l@{}}- 30,000 video frames\\ - Frontal views\end{tabular} & 4 & \begin{tabular}[c]{@{}l@{}}- Controlled\\ - Spontaneous\end{tabular} & \begin{tabular}[c]{@{}l@{}}- Valence\\ - Quantized\\ - Continuous\end{tabular} \\ \hline
RELOCA~\cite{ringeval2013introducing} & \begin{tabular}[c]{@{}l@{}}- Multi-modal\\ - Audio, Video, ECG and EDA\end{tabular} & 46 & \begin{tabular}[c]{@{}l@{}}- Controlled \\ - Spontaneous\end{tabular} & \begin{tabular}[c]{@{}l@{}}- Valence and arousal (continuous)\\ - Self-Assesment\end{tabular} \\ \hline
AM-FED~\cite{mcduff2013affectiva} & - 242 facial videos & 242 & Spontaneous & - 14 AUs \\ \hline
DEAP~\cite{koelstra2012deap} & \begin{tabular}[c]{@{}l@{}}- 40 one-minute videos\\ - EEG signals recorded\end{tabular} & 32 & \begin{tabular}[c]{@{}l@{}}- Controlled\\ - Spontaneous\end{tabular} & \begin{tabular}[c]{@{}l@{}}- Valence and arousal (continuous)\\ - Self-Assesment\end{tabular} \\ \hline
GFT~\cite{girard2017sayette} & - 172,800 video frames & 96 & \begin{tabular}[c]{@{}l@{}}- Controlled\\ - Spontaneous\end{tabular} & \begin{tabular}[c]{@{}l@{}}- 20 AUs\\ - Facial Landmarks\\ - Head Pose\end{tabular} \\ \hline
B4PD~\cite{zhang2014bp4d} & - 368,036 video frames & 41 & \begin{tabular}[c]{@{}l@{}}- Controlled\\ - Spontaneous\end{tabular} & \begin{tabular}[c]{@{}l@{}}- AUs\\ - 2D/3D Facial Landmarks\end{tabular} \\ \hline
B4PD+~\cite{zhang2016multimodal} & - 1,400,000 video frames & 140 & \begin{tabular}[c]{@{}l@{}}- Controlled\\ - Spontaneous\end{tabular} & \begin{tabular}[c]{@{}l@{}}- 34 AUs\\ - Head Pose\\ - 2D/3D/IR Facial Landmarks\end{tabular} \\ \hline
4DFAB~\cite{cheng20174dfab} & - 1,800,000 3D meshes & 180 & \begin{tabular}[c]{@{}l@{}}- Posed\\ - Spontaneous\end{tabular} & \begin{tabular}[c]{@{}l@{}}- 6 Basic Emotions\\ - Facial Landmarks\end{tabular} \\ \hline
AFEW~\cite{dhall2013emotion} & - 1832 videos & 330 & - Wild & - 7 emotion categories \\ \hline
FER-2013~\cite{goodfellow2013challenges} & \begin{tabular}[c]{@{}l@{}}- 35,887 images\\ - Queried from the web\end{tabular} & 35,887 & - Wild & - 7 emotion categories \\ \hline
EmotioNet~\cite{benitez2016emotionet} & \begin{tabular}[c]{@{}l@{}}- 100,00 images annotated manually\\ - 900,000 images annotated automatically\\ - Queried from the web\end{tabular} & 100,000 & - Wild & \begin{tabular}[c]{@{}l@{}}- 12 AUs\\ - 23 emotion categories based on AUs\end{tabular} \\ \hline
Aff-Wild~\cite{zafeiriou2016facial} & - 500 YouTube videos & 500 & - Wild & - Valence and arousal (continuous) \\ \hline
FER-Wild~\cite{mollahosseini2016facial} & \begin{tabular}[c]{@{}l@{}}- 24,000 images\\ - Queried from the web\end{tabular} & 24,000 & - Wild & - 7 emotion categories \\ \hline
AffectNet~\cite{mollahosseini2017affectnet} & \begin{tabular}[c]{@{}l@{}}- 1,000,000 images with facial landmarks\\ - 450,000 images annotated manually\\ - Queried from the web\end{tabular} & 450,000 & - Wild & \begin{tabular}[c]{@{}l@{}}- 8 emotion categories\\ - Valence and arousal (continuous)\end{tabular} \\ \hline
Belfast Database~\cite{sneddon2012belfast} & - 1,400 videos & 256 & - Wild & \begin{tabular}[c]{@{}l@{}}- Disgust, Fear, Anger, Surprise\\ - Frustration\end{tabular} \\ \hline
\textbf{\begin{tabular}[c]{@{}c@{}}DAiSEE\\ (This Work)\end{tabular}} & \textbf{\begin{tabular}[c]{@{}l@{}}- 9,068 video sequences\end{tabular}} & \textbf{112} & \textbf{- Wild} & \textbf{\begin{tabular}[c]{@{}l@{}}- Engagement\\ - Boredom\\ - Confusion\\ - Frustration\end{tabular}} \\ \hline
\end{tabular}%
}
\end{table*}

% As mentioned earlier, until recently, most efforts focused on the seven basic expressions (neutral, happiness, sadness, anger, disgust, fear, surprise) and the facial action units connected with them, as in \cite{yang2007boosting}\cite{tian2001recognizing}. Despite recent efforts to go beyond these basic expressions as well as model affective states in terms of dimensions such as valence and arousal \cite{gunes2011emotion}\cite{he2015multimodal}\cite{valstar2014avec}\cite{rehg2013decoding}, very limited work has been carried out in perceiving affective states related to user engagement, which has direct relevance to various applications described in the earlier section.
% Put more data here

\subsection{User Engagement}
Hernandez et al.~\cite{hernandez2013measuring} were the first to attempt the problem of recognizing user engagement. They modeled the problem of determining engagement of a TV viewer as a binary classification task, using facial geometric features and SVMs for classification. They created a very small custom dataset (which is not publicly available), labeled by a single coder for the presence of engagement. Similarly, Whitehill et al.~\cite{whitehill2014faces} attempted to automatically understand engagement in learning environments. They developed a custom dataset labeled by a few coders, which again is not publicly available. They experimented with feature extraction methods and classifiers and concluded that Support Vector Machines with Gabor features gave the best accuracy on recognizing user engagement on a scale of four levels (highest accuracy being $\approx 69\%$). Besides, the dataset was captured under constrained settings and did not capture the "in-the-wild" needs of real-world user engagement recognition. 

\subsection{Commerical Software}
\begin{figure*}[h]
\centering
\subfloat[High Engagement]{\includegraphics[width=58mm, height=40mm]{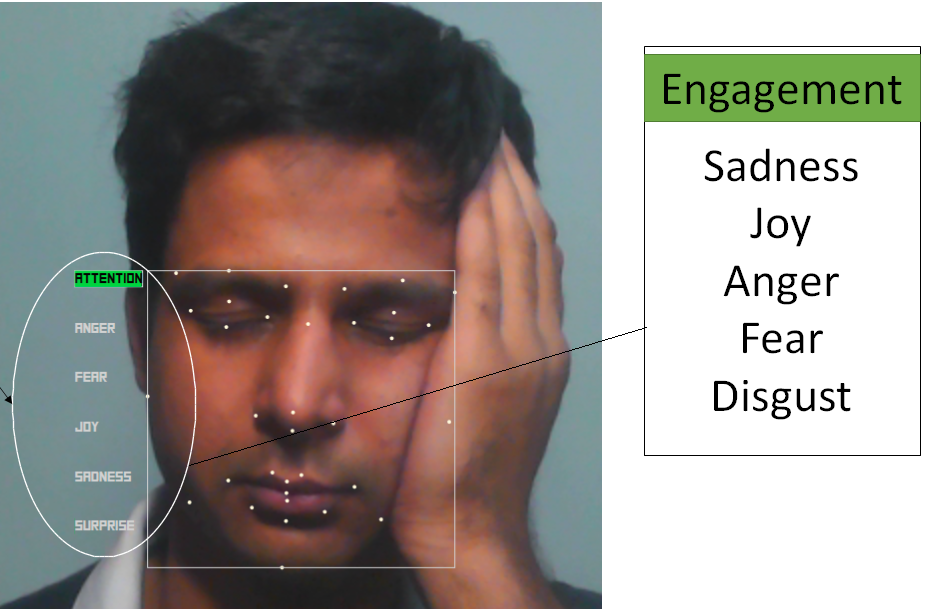}}
\subfloat[Zero Engagement]{\includegraphics[width=58mm, height=40mm]{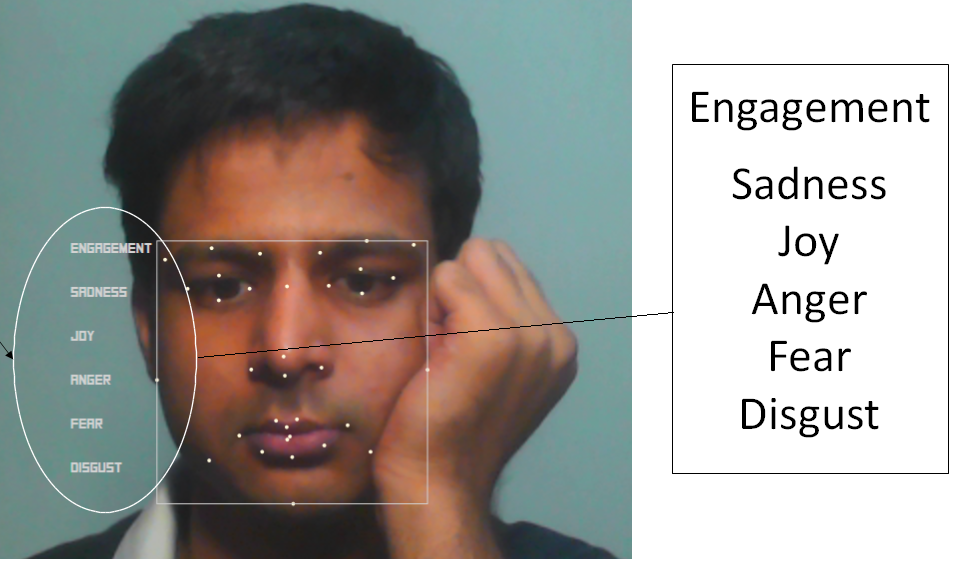}}
\subfloat[Minimal Engagement]{\includegraphics[width=58mm, height=40mm]{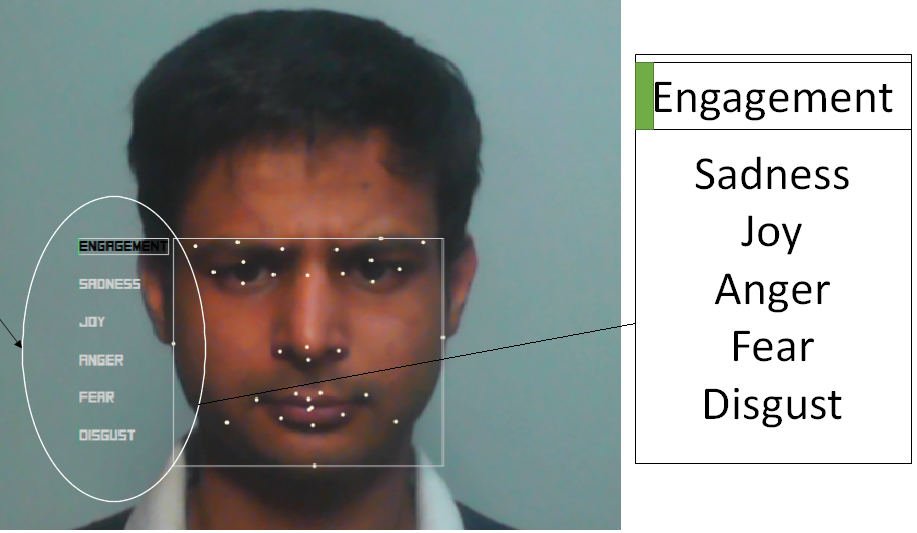}}
    \caption{(Best viewed in color) Results of Affdex on videos from our dataset. The topmost entry on the text-inset in each figure shows the level of engagement. In the image (a), the software detects that the subject is engaged (full green bar), even though the subject is not. In subfigure (b), the software detects zero engagement (zero green bar), even though the subject is engaged. In subfigure (c), the software detects a highly engaged person as having minimal engagement (low green bar). All levels of engagement are determined by psychologists}
\label{fig_commercial}
\end{figure*}

The relevance of affective state recognition in real-world applications can be gauged by the rising number of commercial applications that attempt to address this challenge. Applications such as Emotient~\cite{emotient}, Emovu~\cite{emovu}, and Sightcorp~\cite{sightcorp} provide an estimation of comparable affective states (called attentiveness, for instance, in SightCorp) in their frameworks. All these applications are firstly constrained only to attentiveness/engagement and do not consider other related affective states, such as boredom or confusion (which we seek to capture in this work). More importantly, our studies with these applications showed that their performance on real-world videos is far below satisfactory, thus highlighting the need for a dataset that captures real-world conditions for further research. Figure~\ref{fig_commercial} shows an example of the performance of Affdex~\cite{affdex} on videos from our dataset and we see that the software shows a user to be attentive even if the user's eyes are closed or the user is looking away from the screen. We note from the image that the software tracks facial key points and correlates them with emotional and cognitive states. Other applications such as SightCorp ~\cite{sightcorp} use the eye gaze of the subject as the sole determinant of the engagement level. While these are good starting points, most of them re-purpose existing methods to detect engagement, and the availability of a larger dataset will promote better methods for reliable engagement recognition. 

\subsection{Why E-Learning?}
Subject to the exponential growth of MOOCs over the last few years, e-learning has received significant attention from several research groups. While machine learning methods have been used to personalize educational modules, diversify assessment methods and make personalized recommendations based on learner preferences and browsing patterns \cite{castro2007applying}\cite{huang2007constructing}\cite{baylari2009design}\cite{pivec2006eye}, limited efforts have been attempted in understanding user engagement in the e-learning environment. Understanding how engaged a student is, is an important task that can increase the learning intake of a student. More importantly, e-learning provides a well-defined context for developing this dataset, as well as studying the impact of effective recognition methods, thereby motivating us to choose this context for this dataset, without any loss of generality. This dataset, however, is not restricted to this domain and can be easily extended to advertising, e-shopping, health-care and autonomous driving to name a few. \\

We believe DAiSEE fulfills this need in the community. In particular, it provides a large enough dataset to train state-of-the-art deep learning models. It provides a significant jump in the nature of affective states and the number of video frames (2,723,882 video frames from 9068 snippets and a total of 25 hours of video recording) from other datasets in this area, as described further in Section \ref{sec_daisee_dataset}.

\section{The Daisee Dataset}
\label{sec_daisee_dataset}

We now present the DAiSEE dataset that contains video recordings of subjects in an e-learning environment, annotated with crowdsourced labels for engagement, frustration, confusion, and boredom. The dataset captures ``in the wild" settings typically seen in the real-world, and will be made publicly available, along with the individual annotations of the crowd, to facilitate open research. We will now discuss the data collection, annotation, and vote aggregation strategies used to create the dataset.

\subsection{Data Collection}
\label{sec_datacollect}
%\subsubsection{Data Capture} 
To model real-life settings, we used a full HD web camera (1920x1080, 30 fps, focal length 3.6mm, 78$^{\circ}$ field of view) mounted on a computer focusing on student users watching videos was used. To simulate the e-learning environment, a custom
application was created that presented a subject with 2 different videos (20 minutes total in length), one educational and one recreational to capture both focused and relaxed settings, which allow natural variations in user’s engagement levels. To model unconstrained settings, the subjects had the option to scroll through the videos.

There are 112 subjects in the dataset belonging to the age group of 18-30, all of whom are currently enrolled students. The race of the subjects is Asian, with 32 female and 80 male subjects. A total of 12583 video snippets are collected, each 10 seconds long. This duration is chosen as Jacob Whitehill \cite{whitehill2014faces} observed that 10-second labeling tasks are more intuitive. After data cleaning, we end up with a dataset that has 9068 video snippets varying across 6 different locations such as dorm rooms, crowded lab spaces, library etc and 3 different illumination settings (light, dark and neutral); with a male-to-female ratio of 2.13:1. Each video snippet is given a unique identification number to help differentiate between the settings of video snippets.

The Hawthorne effect~\cite{mccarney2007hawthorne}~\cite{monahan2010benefits} also referred to as the observer effect, is a type of reactivity where individuals modify an aspect of their behavior in response to their awareness of being observed. This is a critical aspect of such a data capture setting and it is highly probable that the subjects may adapt their behavior to suit the objectives of the experiment. To limit the occurrence of such circumstances, the subjects are recorded without being trained or setting any parameters for the experiment. 

\subsubsection{Subject Privacy}
All participants who appear in the video snippets have given signed consent for the recorded videos to be distributed for use by the wider research community. In the event consent is declined, the captured videos are deleted. Further, the anonymity of every subject is ensured by giving him/her a uniquely generated 3-digit id whose correspondence with the identity is not recorded anywhere.

% To account for subject privacy, they are informed of the recordings and consent is taken to present their work. In the event consent is declined, the captured videos are removed. Further, the anonymity of every subject is ensured by giving him/her a uniquely generated 3-digit id whose correspondence with the identity is not recorded anywhere.

\subsection{Data Annotation}
Motivated by recent work in intelligent tutoring systems~\cite{rajendran2014enriching}, our dataset consists of labels for four affective states related to user engagement, viz., engagement, frustration, confusion, and boredom. Recent work~\cite{calvo2010affect} has shown that the six basic expressions: anger, disgust, fear, joy, sadness, and surprise~\cite{eckman1972universal} are not reliable in prolonged learning situations, as they are prone to rapid changes. Each of the affective states is defined at four levels: (1) very low (2) low (3) high and (4) very high (similar to \cite{whitehill2014faces}). We followed this labeling strategy to avoid the ``neutral" state since early experiments showed that crowd annotators often preferred to choose ``neutral"  as a state when unsure. The possible levels for labels ensured that the annotators ``took a stand" on the affective state level, which is essential for a robust dataset. We also note that as in \cite{whitehill2014faces}, the above 4-level annotation can be trivially changed to 2-levels (\textit{high} and \textit{low}) when required for a given application setting.

Subtle affective states such as user engagement are subjective and vary based on the viewer's discretion. Hence, we rely on ``wisdom-of-the-crowd" for our annotations in the dataset. The easy availability of annotators on crowdsourcing platforms has resulted in many new computer vision datasets such as \cite{deng2009imagenet, lin2014microsoft, van2015building} which tap into the crowd for annotating large data. Although the annotators can be non-experts, it has been shown that repeated labeling of examples by multiple annotators produces high-quality labels~\cite{sheng2008get, welinder2010online, ipeirotis2014repeated}. In this work, we used CrowdFlower, for the annotations, similar to~\cite{burke2011crowdsourcing}. CrowdFlower provides advanced quality control mechanisms, worker targeting and detailed reports on the final annotation results obtained. Other features of CrowdFlower that we used in this work are mitigation of bot labeling, priming of annotator to the specific task using reasoned test questions, and flagging labels of underperforming annotators.

To obtain the votes, each annotator is presented with a video snippet and asked to vote. Annotators are presented with instructions on how to perform the task and illustrative examples to facilitate the process. Additionally, each annotator answers a standardized test question, that helps us remove the votes of under-performing annotators. For each video snippet, we get votes from 10 different annotators (which is comparable to other standard crowdsourced datasets such as \cite{zhou2012learning}\cite{han2015demographic}).

\subsection{Vote Aggregation}
Vote aggregation is used to assign a label for each affective state in a given video snippet, using the annotations from the crowd. We use the Dawid-Skene~\cite{dawidskene} vote aggregation algorithm to obtain the ground truth label for each snippet, since this is often considered `gold standard' for aggregation in practice. Dawid-Skene is an unsupervised inference algorithm that gives the Maximum Likelihood Estimate of observer error rates using the EM algorithm. However, before vote aggregation, we removed faulty annotations and noise in order to increase the robustness of votes and reliability of annotators.

\subsubsection{Removing Bad Annotators}
The annotators on CrowdFlower are not experts and noisy annotations cause the aggregation algorithm to output false labels for a video snippet. DAiSEE has a total of 1690 unique annotators who voted on 12583 video snippets, with each snippet having 10 annotations. To remove noisy labels, we created a gold standard for a subset of 1157 unique video snippets, which included every annotator. Two teams of experts, each consisting of a social, a clinical and a behavioral psychologist from the department of psychology at our institute were formed. Each group worked on mutually exclusive videos from the subset and consensus from each of the three experts used to create the gold standard. This is used to identify and remove faulty annotators.

To estimate an annotator's reliability, for each set of annotations, we use the gold standard to establish the inter-coder agreement. Since the labels for any video snippet are ordinal ({1,2,3,4}) in nature, we use a weighted Cohen's $\kappa$ (score between 0-1) with quadratic weights to penalize label disagreement between the annotator and the gold standard. Any annotator whose agreement coefficient is less than 0.5 is marked as unreliable, their annotations as noisy, and removed from the dataset.

% weighted kappa measures and table to be put here

The resulting dataset has video snippets with annotations varying from 0-10 in number, with a median of 4 annotations. All video snippets with less than 4 annotations are removed, resulting in a dataset with 9,068 snippets and 2,723,882 frames.

\subsubsection{Dawid-Skene Aggregation}
We use the Dawid-Skene~\cite{dawidskene} vote aggregation strategy to obtain the ground truth label for each snippet. Dawid-Skene is an unsupervised inference algorithm that gives the Maximum Likelihood Estimate of observer error rates using the EM algorithm.

\begin{enumerate}
\setlength\itemsep{0em}
\item Using the labels given by multiple annotators, estimate the most likely ``correct" label for each video snippet.
\item Based on the estimated correct answer for each object, compute the error rates for each annotator.
\item Taking into consideration the error rates for each annotator, recompute the most likely ``correct" label for each object.
\item Repeat steps 2 and 3 until one of the termination criteria is met (error rates are below a pre-specified threshold or a pre-specified number of iterations are completed).
\end{enumerate}

The distribution of labels after removing bad annotators is depicted in Table~\ref{tab_daisee_labels}

\begin{table}[h]
\caption{Distribution of labels in DAiSEE across its affective states}
\label{tab_daisee_labels}
\centering
\begin{tabular}{|c|c|c|c|c|}
\hline
\textbf{Affective State} & \textbf{Very Low} & \textbf{Low}  & \textbf{High} & \textbf{Very High} \\ \hline
Boredom         & 3869     & 2931 & 1934 & 334       \\ \hline
Confusion       & 6024     & 2191 & 752  & 101       \\ \hline
Engagement      & 61       & 459  & 4477 & 4071      \\ \hline
Frustration     & 6986     & 1649 & 346  & 87        \\ \hline
\end{tabular}
\end{table}

\subsection{How good is DAiSEE?}

\begin{figure}[h]
\centering
\includegraphics[width=87mm, height=25mm]{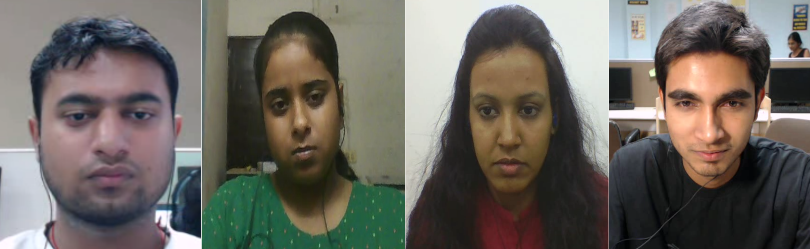}
\caption{From left to right we can see engagement vary from very low to very high}
\label{fig_level_engagement}
\end{figure}

\begin{figure}[h]
\centering
\includegraphics[width=87mm, height=25mm]{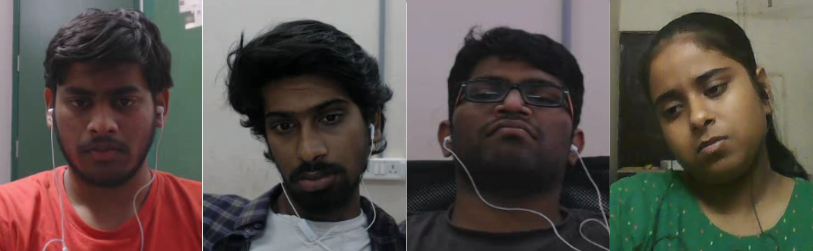}
\caption{From left to right we can see boredom vary from very low to very high}
\label{fig_level_boredom}
\end{figure}

\begin{figure}[h]
\centering
\includegraphics[width=87mm, height=25mm]{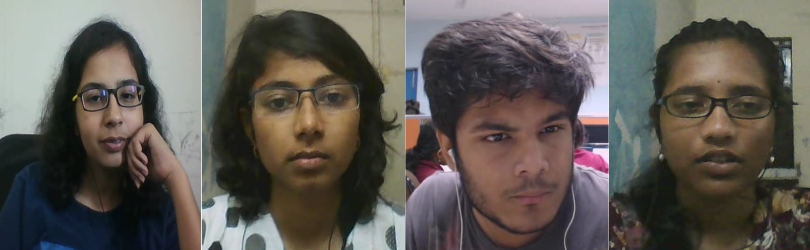}
\caption{From left to right we can see confusion vary from very low to very high}
\label{fig_level_confusion}
\end{figure}

\begin{figure}[h]
\centering
\includegraphics[width=87mm, height=25mm]{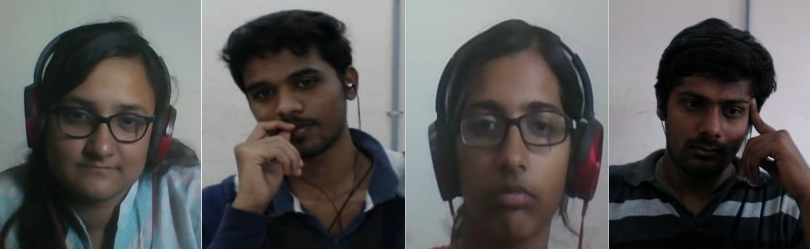}
\caption{From left to right we can see frustration vary from very low to very high}
\label{fig_level_frustration}
\end{figure}

Examples of different levels of each affective state is depicted in Figures~\ref{fig_level_engagement}, ~\ref{fig_level_boredom}, ~\ref{fig_level_confusion} and ~\ref{fig_level_frustration}. 

\begin{figure}[h]
\centering
\includegraphics[width=87mm, height=25mm]{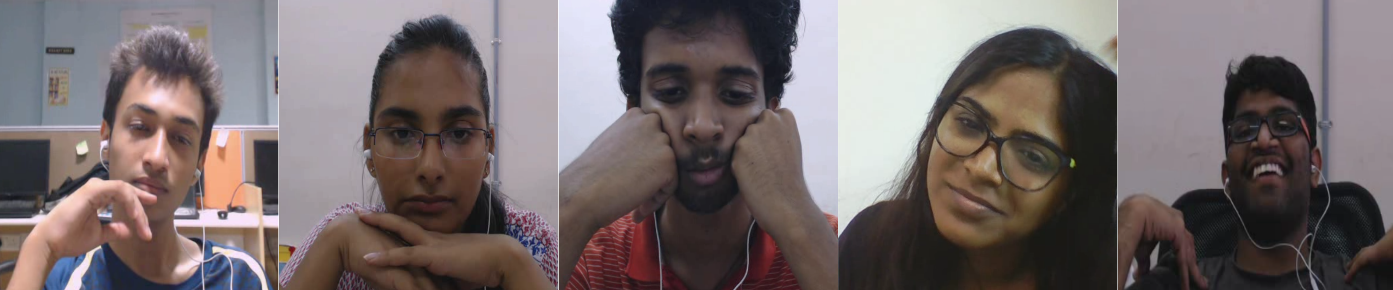}
\caption{Variety in DAiSEE : All subjects have Very High Engagement but express it through various poses. Illumination changes are visible from light to dark in the above images as we move from left to right}
\label{poses}
\end{figure}

The ``in the wild'' settings of DAiSEE can be seen in Figure~\ref{poses} where we see how different illumination settings change across images and how different people express very high engagement through different poses.

DAiSEE contains high-fidelity annotations for all videos. To establish this, we compared the labels obtained from Dawid-Skene after removal of bad annotators and the labels from experts on the gold standard and observe that for 70\% of the videos, the labels match with the gold standard. To further study the correlation of DAiSEE's final aggregated labels with layman users, we conducted a user study with 10 randomly chosen subjects on 100 videos and asked each user to agree or disagree with the label associated with a video snippet. The users were not given any further information about the nature or background of this work. Our results of this study showed that on 84 of the 100 videos, the majority aggregation of the users' opinions agreed with the aggregated labels from the crowd. Among the remaining 16 videos, based on subjective feedback gathered, we found that the disagreement was about the level of intensity of the label (low vs very low engagement), rather than of labels with opposite polarities (very high vs very low engagement) in 10 videos, and of the affective state with opposite polarity in 6 videos.

The unique features of DAiSEE are summarized below:

\begin{itemize}
\item It is the first publicly available dataset for studying user engagement and related affective states (in the wild).
\item While there has been substantial work in recognition of the seven basic emotions, DAiSEE presents a dataset to understand more subtle affective states, such as engagement and boredom, which are often not exhibited explicitly on a human. We believe that this will help advance the field in a new direction, with potential applications in several fields (as described earlier).
\item It contains videos, thus allowing researchers to use the temporal information for effective recognition.
\item DAiSEE will be publicly available and will include both raw crowd annotations, as well as the high-fidelity aggregated annotations obtained through the process described above, for all video snippets. The crowd annotations could be used to develop better vote aggregation methods.
% \item It is the largest dataset for affective state recognition/detection and is at least three times larger than any existing facial expression dataset.
\end{itemize}

DAiSEE presents a platform for advancing applications related to e-learning, healthcare, advertising and autonomous vehicles which benefit from recognition of user engagement and associated affective states. We now present the benchmark results we obtained on DAiSEE using many state-of-the-art video classification methods, as described in Section \ref{sec_benchmark}.

\section{Benchmark Results}
\label{sec_benchmark}
Majority of the videos in DAiSEE are captured with the frame of capture extending up to the bust of the user. This allows interested researchers to study the relevance of non-facial cues such as upper body postures on user engagement (for example, a laidback posture could indicate higher levels of boredom, while an upright posture could indicate high levels of engagement). We use a number of state-of-the-art deep learning models in order to assess how well they are working on DAiSEE. We run experiments using two types of models; \emph{Static} (Frame classification/prediction) and \emph{Dynamic} (temporal video classification). In order to be model-agnostic while experimenting, minimal hyper-parameter tuning is done. All experimental results are based on vanilla implementations of models. We now discuss the models, their performance on DAiSEE and the final benchmarking numbers.

\subsection{Dataset Split}
To prepare the dataset for benchmarking, we create a data-split into train, validation and test sets. The following three principles are used:
\begin{itemize}[noitemsep,topsep=0pt]
\item For training deep learning models, follow the general Kaggle \cite{kagsplit} practice of 60:20:20 for the train:validation:test sets
\item All splits are mutually exclusive and exhaustive with respect to subjects
\item The same ratio of males:females is maintained across the splits
\end{itemize}

\subsection{Models}
We experiment with two types of models, temporal and static. All models are based on Convolutional Neural Networks (CNNs) \cite{cnn@karp} since they have shown ground-breaking results in the area of computer vision in the past years. Since we used vanilla implementations with minimal hyper-parameter tuning, we used bi-linear interpolation to re-shape the image and fed it to the model. Each affective state is benchmarked individually using the given models:

\begin{itemize}[noitemsep,topsep=0pt,leftmargin=*]
\item \textbf{Single Frame Classification}: We did frame level classification and accuracy measurement. We used a InceptionNet V3 model, pre-trained using ImageNet\cite{imageNet@static}; restricting the output softmax layer to 4. We fine-tuned the top dense layers and then trained the top two inception blocks. The images were reshaped to (229x229x3) and the data splits shuffled to not introduce sequencing before being used as input to the model. 

\item \textbf{Single Frame Prediction, Video Classification}: We did frame level predictions and video level accuracy measurements. No shuffling is done on the input, and all the frames for a given snippet are fed to the model, their outputs aggregated for a video snippet and measured for accuracy. We used the same model as single frame classification. This is done to observe if temporal models are needed or pooled results from static networks would be sufficient.

\item \textbf{Video Classification, Full training}: We use C3D \cite{c3D@fair} which is a modified version of BVLC Caffe \cite{caffe} to train a 3D CNN. The model is fully trained using our data splits, an output softmax layer for four classes and accuracy measured against the label for a video.

\item \textbf{Video Classification, Transfer learning}: We use the pre-trained C3D model on Sports-1M \cite{cnn@karp}, and fine-tuned it for our dataset by modifying the output softmax layer for four classes and measuring accuracy against the label for a video.

\item \textbf{Video Classification, Sequence learning}: We used a Long-Term Recurrent Convolutional Network \cite{donahue2015long} for unifying visual and sequence learning. We did two modifications: 1) As subjects are mostly static, optical flow does not provide any useful information and hence it is not computed; 2) We recreated the video by considering every alternate frame as continual affective states such as engagement do not vary in less than 30 ms. After modifying the output softmax layer for four classes, we did a full training for our data splits and measured accuracy against the label for a video.

\end{itemize}

\subsection{Performance Metrics}
All the classification models were tested on Top-1 accuracy. We ran each model described in the previous section 3 times to avoid any randomness bias (due to factors such as weight initialization in the models). We present our results and analysis of the dataset below.

\subsection{Results}
\label{subsec_results}
\begin{figure}[h]
\begin{center}
\includegraphics[width=87.7mm, height=60mm]{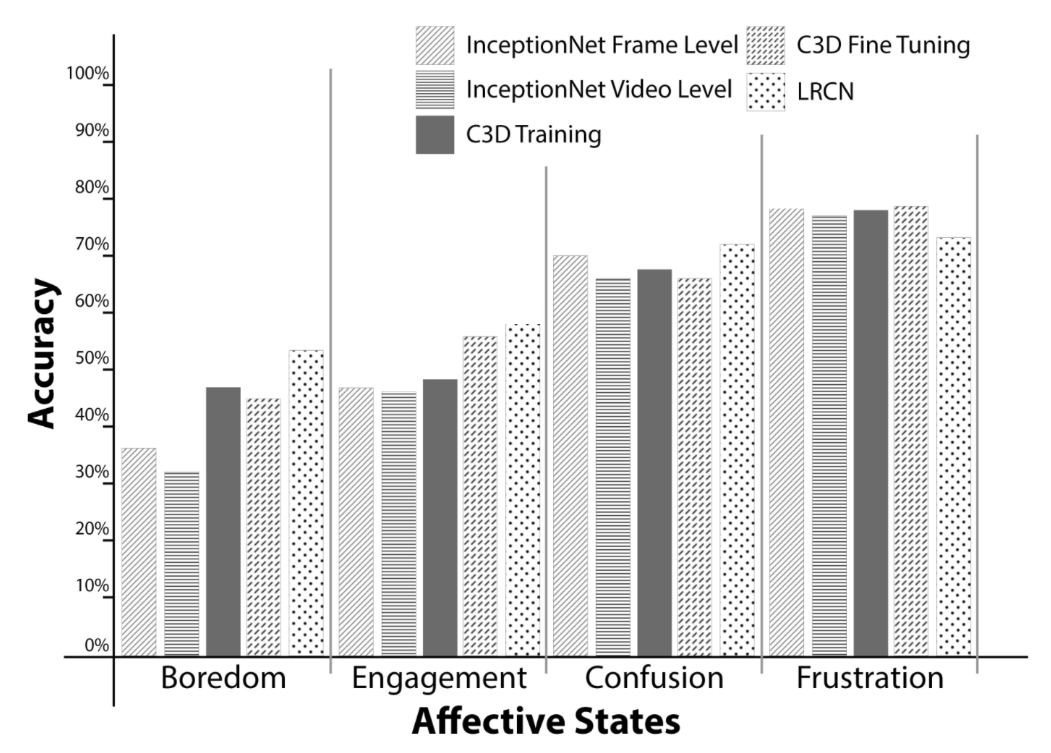}
\end{center}
\caption{Results of various experiments on DAiSEE. We report top-1 accuracy with five-fold cross-validation}
\label{results}
\end{figure}

Figure~\ref{results} and Table~\ref{tab_results_models} show the baseline accuracy results from our studies. We see that LRCN generally performs better than all other models. Additionally, the experiments also show that temporal classifiers tend to outperform the static classifiers illustrating that affective states such as engagement, frustration, boredom, and confusion are prolonged in nature and cannot be estimated by looking at a particular instant in time. Experiment on InceptionNet with frame level prediction/video classification showed us that the accuracy is only slightly better than random guessing for boredom, again illustrating that static models are insufficient for analyzing such affective states.

\begin{table*}[h]
\caption{Benchmark results on DAiSEE. We report Top-1 Accuracy averaged over three runs}
\label{tab_results_models}
\centering
\resizebox{\textwidth}{!}{%
\begin{tabular}{|c|c|c|c|c|c|}
\hline
\textbf{Affective State} & \textbf{InceptionNet Frame Level} & \textbf{InceptionNet Video Level} & \textbf{C3D Training} & \textbf{C3D FineTuning} & \textbf{LRCN} \\ \hline
Boredom & 36.5\% & 32.3\% & 47.2\% & 45.2\% & 53.7\% \\ \hline
Engagement & 47.1\% & 46.4\% & 48.6\% &  56.1\% & 57.9\% \\ \hline
Confusion & 70.3\% & 66.3\% & 67.9\% & 66.3\% & 72.3\% \\ \hline
Frustration & 78.3\% & 77.3\% & 78.3\% & 79.1\% & 73.5\% \\ \hline
\end{tabular}}
\end{table*}

\subsection{Analysis}
\label{subsec_analysis}

% We performed additional analysis on the dataset to 

\textbf{Effect of removing bad annotators}: We did a study to see how many labels improved (came closer to the true level) for the gold standard for each affective state after the removal of bad annotators. We followed a two-step approach to i) run Dawid-Skene without removal of bad annotators and ii) run Dawid-Skene with the removal of bad annotators and then compare them against the annotations by the experts (psychologists) to test for any improvement in label quality. The results are summarized in Table~\ref{tab_annotator_removal}.

\begin{table}[h]
\caption{Improvement in label quality after removal of bad annotators for different affective states of DAiSEE in the gold standard}
\label{tab_annotator_removal}
\centering
\begin{tabular}{|c|c|}
\hline
\textbf{Affective State} & \textbf{\% of Modified Labels} \\ \hline
Engagement & 5.78 \\ \hline
Boredom & 2.46 \\ \hline
Confusion & 10.82 \\ \hline
Frustration & 8.12 \\ \hline
\end{tabular}
\end{table}

\textbf{Using EmotionNet to analyze DAiSEE}: To further analyze DAiSEE, we use EmotionNet~\cite{benitez2016emotionet}, a CNN used for emotion recognition in photographs of human faces as a benchmarking model. We used the implementation of EmotionNet given by authors~\footnote{https://github.com/co60ca/EmotionNet2}. EmotionNet is pre-trained using CK+~\cite{lucey2010extended} and Karolinska Directed Emotional Faces (KDEF)~\cite{goeleven2008karolinska} datasets. Before passing the video as input to the model, we used FaceNet \cite{schroff2015facenet} to detect and extract faces. In the event that a face was not detected, we used LabelImg~\cite{labelimg} to manually extract the face. We, then, fine-tune the model on DAiSEE by sampling every fourth frame from the video (Our experiments show that this does not affect accuracy). The results are summarized in Table~\ref{tab_emotionet}

\begin{table}[h]
\caption{Benchmarking EmotionNet on DAiSEE}
\label{tab_emotionet}
\centering
\begin{tabular}{|c|c|}
\hline
\textbf{Affective State} & \textbf{Accuracy} \\ \hline
Engagement & 51.07\% \\ \hline
Boredom & 35.89\% \\ \hline
Confusion & 57.45\% \\ \hline
Frustration & 73.09\% \\ \hline
\end{tabular}
\end{table}

\begin{figure}[h]
\centering
\subfloat[Low Engagement]{\includegraphics[width=30mm, height=30mm]{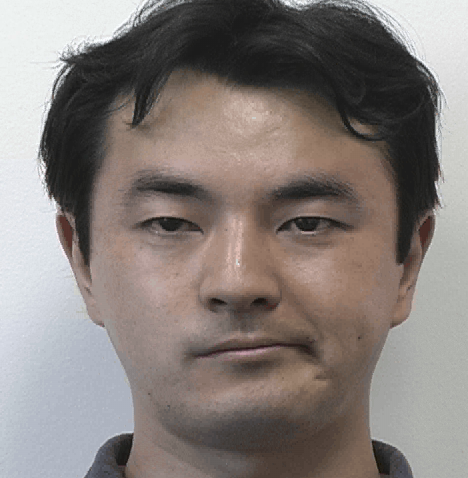}} 
\subfloat[High Engagement]{\includegraphics[width=30mm, height=30mm]{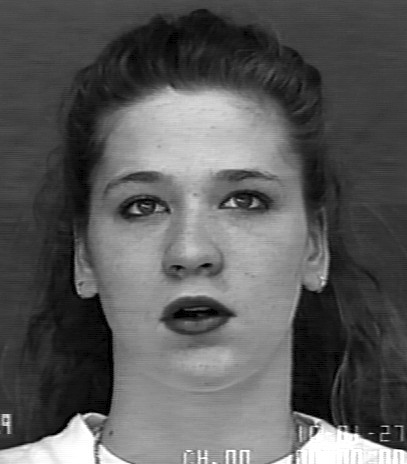}} 
\subfloat[Very-High Engagement]{\includegraphics[width=30mm, height=30mm]{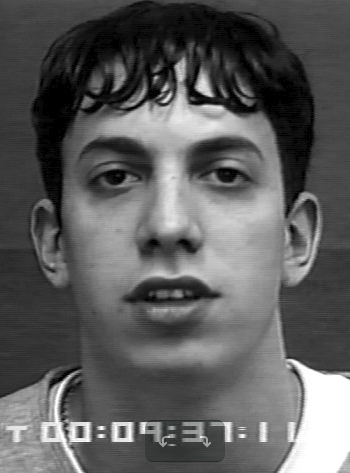}} 
\caption{Samples of pre-trained EmotionNet on CK+. In the first image, we see ``low'' level of engagement. In the second image, we see ``high'' level of engagement and in the third image we see ``very-high'' level of engagement}
\label{fig_transfer_testing}
\end{figure}

\textbf{Using pre-trained EmotionNet model on CK+}: We use the EmotionNet model trained for engagement and pass various images from the CK+~\cite{lucey2010extended} database. For each image, we obtain a label corresponding to the level of engagement the network predicts. Some sample outputs are shown in Figure~\ref{fig_transfer_testing}. The model does not output ``very-low'' engagement for any image. We believe that the reason for this is because the subjects in the CK+ database are well-aware of their environment (the experiments are controlled) and hence are engaged during the course of the experiment.

\textbf{Considering each label in a binary classification setting}: We also considered the problem of binary classification for engagement, by trivially changing the labels from (low, very low) categories to 'not-engaged' and from (high, very high) categories to 'engaged'. This study was done by using multiple subsets of the data (identical in size), with the same ratio of engaged:not-engaged labels as DAiSEE and averaging the results of these multiple subsets. Using LRCN, we achieved a Top-1 accuracy of 94.6\%. This shows the challenges when trying to learn the subjective levels for engagement compared to learning whether the person is engaged or not and provides directions for further research in improving techniques to recognize multiple levels for any affective state.

\section{Benchmarking Challenges of Daisee}
\label{sec_benchmarking_challenges}

\begin{figure}[h]
\centering
\includegraphics[width=25mm, height=30mm]{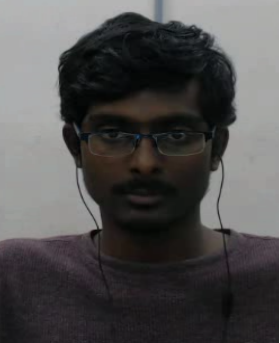}
\caption{This figure depicts an instance of low illumination in DAiSEE. The low illumination makes it difficult for the faces to be detected and hence the model cannot learn appropriate features required for correct classification}
\label{fig_low_illum}
\end{figure}

While DAiSEE shows significant numbers for accuracy, the ``in-the-wild'' settings of DAiSEE provide some significant challenges in benchmarking the dataset, resulting in possible misclassification of videos. From our analysis, such misclassified videos are shown to have at least one of the following characteristics:

\textbf{Low Illumination}: One of the most common problems was that of poor illumination. In videos with low illumination, the faces are not clearly visible, making it difficult to capture the affective state of the individual, as shown in Figure~\ref{fig_low_illum}.

\begin{figure}
\begin{center}
\includegraphics[width=87mm, height=23mm]{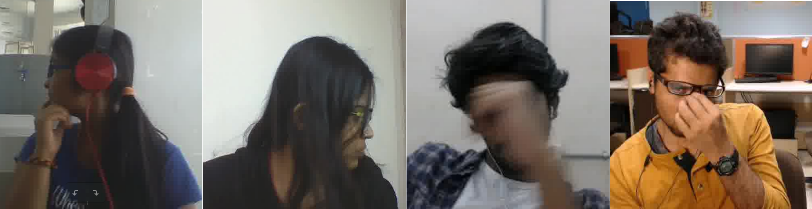}
\end{center}
\caption{Examples displaying the lack of frontal face alignment or facial occlusion}
\label{fig_faceoccalign}
\end{figure}

\textbf{Face Occlusion/Lack of Frontal Pose}: Another cause for misclassification is when there is occlusion, for example, the hair of a female user covering her face or a user using his/her hands to cover parts of the face. Also, since DAiSEE is captured in the wild, distractions such as opening doors or people walking by caused the subject to turn around, hiding frontal facial features, thus resulting in misclassification. Some samples can be seen in Figure~\ref{fig_faceoccalign}.

\begin{figure}
\begin{center}
\includegraphics[width=87mm, height=23mm]{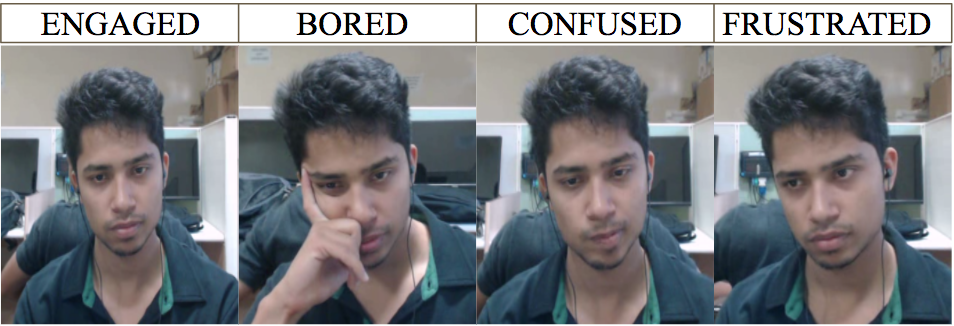}
\end{center}
\caption{Example of the range of affective states observed in a single 10 second video snippet}
\label{changes}
\end{figure}

\textbf{Variety of Affective States Displayed}: There are several video snippets where the subject displays a change in affective states expressed as the snippet progresses, making it difficult for the model to learn essential features to classify the video correctly. An example of such a video can be seen in Figure~\ref{changes}.

% \begin{figure}
% \begin{center}
% \includegraphics[width=87mm, height=25mm]{variety.png}
% \end{center}
% \caption{Another example of the range of affective states observed in a single 10 second video snippet}
% \label{changes2}
% \end{figure}

\begin{figure}[h]
\centering
\subfloat[Highly Bored]{\includegraphics[width=25mm, height=25mm]{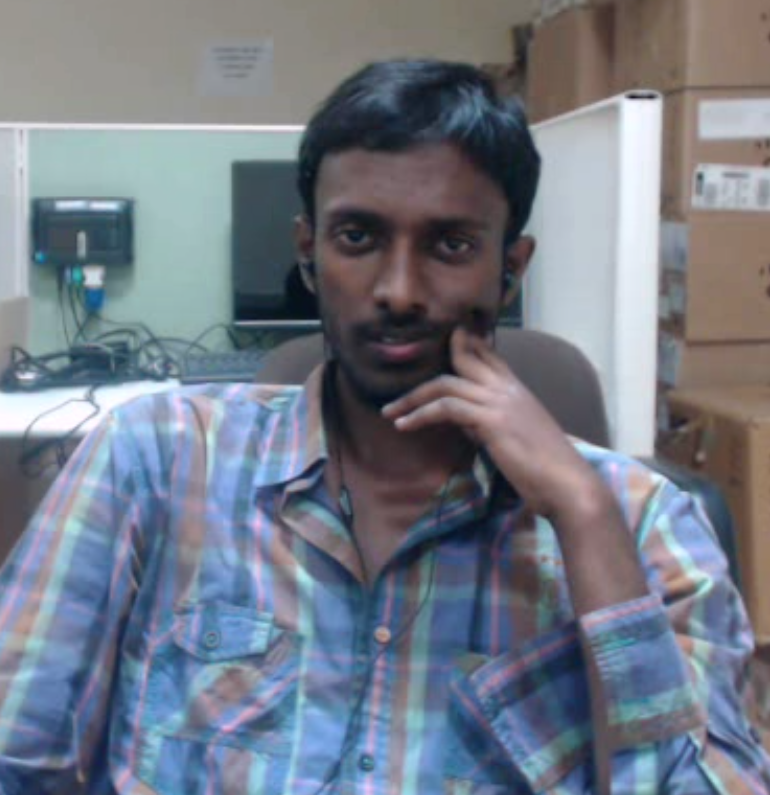}} \hspace{10mm}
\subfloat[Highly Engaged]{\includegraphics[width=25mm, height=25mm]{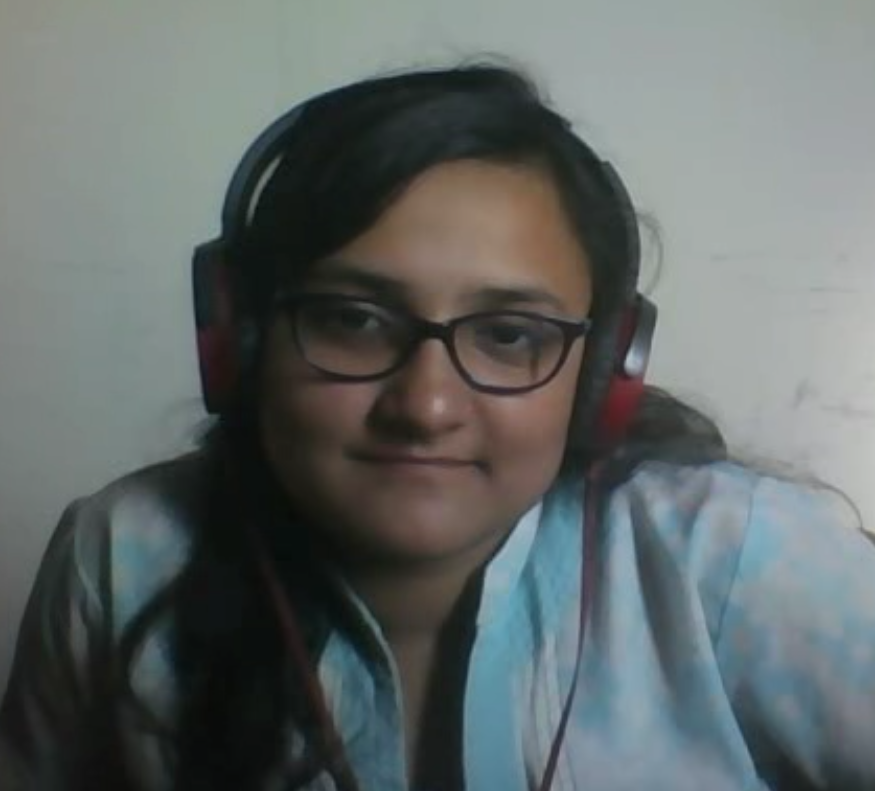}}
\caption{From the figure, we can see that boredom and engagement are complementary in nature to each other. In figure (a) we observe high boredom, but low engagement whereas in figure (b) we observe high engagement, but low boredom}
\label{fig_comp_label}
\end{figure}

\begin{figure}[h]
\centering
\includegraphics[width=87mm, height=25mm]{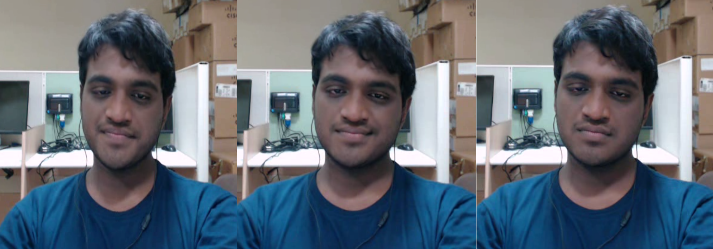}
\caption{In the figure we can see that boredom and engagement are not complementary in nature to each other. The labels for boredom and engagement are 0 while that of frustration is 2. This clearly shows that there is no direct correlation between engagement and boredom}
\label{fig_not_comp_label}
\end{figure}

\textbf{Complementary Labels}: While working with the dataset, we noticed how boredom and engagement complimented each other. When engagement is low, boredom is generally high and vice-versa. In the instances that both boredom and engagement were low for a video snippet, the subject displayed high levels of confusion or frustration. This shows the subtleness but disparate nature of different affective states and how well DAiSEE captures these changes. This is another direction for future research where the correlation between labels is used to develop robust models that identify these affective states simultaneously. We can see in Figure~\ref{fig_comp_label} that boredom and engagement are complementary to each other. On the other hand, the two affective states are not complementary in Figure~\ref{fig_not_comp_label}.

\section{Daisee Release}
\label{sec_release}

DAiSEE is available for download  on \url{https://people.iith.ac.in/vineethnb/resources/daisee/index.html}.

\section{Conclusion and Future Work}
\label{sec_conc}
In this work, we present DAiSEE, a dataset for user engagement in the wild. The novelty of DAiSEE comes from the rich information that it has of different affective states such as engagement, boredom, confusion, and frustration. Each affective state is categorized from very low to very high (without neutral) and is annotated using ``wisdom-of-the-crowd''. The dataset captures the nature of real-world e-learning environments in an organic manner, with varying user poses, positions and background noises typically observed in such settings. It is unique as this is the first publicly available dataset to study these four affective states compared to the seven emotion categories and is the largest available facial emotion/expression dataset for the research community to work on. We also present benchmarking results for DAiSEE and establish a baseline for the research community to build on. To help create a more open community for DAiSEE, we present its raw annotation data for conducting research to improve vote aggregation algorithms or for using these annotations in the training process to improve upon the baselines shared in this work.

Going forward, methods that determine geometric features (such as facial fiducials), facial action units, body and head pose; gaze and gesture can be used as input to models that learn to recognize user engagement. Also, systems that use such mid-level cues, such as pose and gesture, can often be used to develop cognitive models of the subject which include engagement, attentional focus, and intention.
 
We hope that DAiSEE assists teachers, content creators and students in the domain of e-learning, advertisement makers, medical professionals and autonomous vehicle companies in creating better and more responsive systems to help improve human-computer interaction.

\ifCLASSOPTIONcaptionsoff
  \newpage
\fi



%\bibliographystyle{IEEEtran}

% \end{thebibliography}

\bibliographystyle{IEEEtran}
\bibliography{IEEEabrv,egbib}


% \begin{IEEEbiography}{Abhay Gupta}
% \end{IEEEbiography}

% \begin{IEEEbiographynophoto}{Abhay Gupta}

% \begin{IEEEbiographynophoto}{Arjun D'Cunha}

% % insert where needed to balance the two columns on the last page with
% %\newpage

% \begin{IEEEbiographynophoto}{Kamal Awasthi}

% \begin{IEEEbiographynophoto}{Vineeth N Balasubramanian}


%\vfill


\end{document}